\documentclass{article}
\usepackage{spconf}
\usepackage[hidelinks]{hyperref}

\usepackage{iftex} 
\ifLuaTeX
  \usepackage[no-math]{fontspec}
  \usepackage{graphicx}
\else
  \usepackage[dvipdfmx]{graphicx} 
\fi

\usepackage{caption} 
\usepackage{subfig} 

\usepackage{citesort}
\usepackage{stmaryrd, amsmath}
\usepackage{ascmac}
\usepackage{cite}
\usepackage{amsbsy}
\usepackage{amssymb}
\usepackage{latexsym}
\usepackage{xcolor}
\usepackage{empheq}

\usepackage{algorithm}
\usepackage{algorithmicx, algpseudocode}

\makeatletter
\renewcommand{\ALG@beginalgorithmic}{\footnotesize}
\makeatother

\usepackage{bm}
\usepackage[utf8]{inputenc}
\usepackage{breqn}
\graphicspath{{figs/}}

\usepackage{mathtools}
\usepackage{enumitem}
\usepackage{amsthm}
\usepackage{mathrsfs}

\newtheorem{proposition}{Proposition}

\newtheorem{theorem}{Theorem}

\newtheorem{fact}{Fact}
\newtheorem{remark}{Remark}

\newcommand{\dom}{\mathop{\rm dom}\nolimits}

\newcommand{\sProx}{\mathop{\rm s\mbox{-}Prox}\nolimits}

\newcommand{\soft}{\mathop{\rm soft}\nolimits}
\newcommand{\Soft}{\mathop{\rm Soft}\nolimits}

\newcommand{\sign}{\mathop{\rm sgn}\nolimits}

\newcommand{\Id}{\mathop{\rm Id}\nolimits}
\newcommand{\range}{\mathop{\rm range}\nolimits}

\newcommand{\gra}{\mathop{\rm gra}\nolimits}

\newcommand{\diag}{\mathop{\rm diag}\nolimits}

\DeclareMathOperator*{\argmin}{arg\,min}

\usepackage[mathlines, switch]{lineno}

\newcommand*\patchAmsMathEnvironmentForLineno[1]{%
  \expandafter\let\csname old#1\expandafter\endcsname\csname #1\endcsname
  \expandafter\let\csname oldend#1\expandafter\endcsname\csname end#1\endcsname
  \renewenvironment{#1}%
  {\linenomath\csname old#1\endcsname}%
  {\csname oldend#1\endcsname\endlinenomath}}%
\newcommand*\patchBothAmsMathEnvironmentsForLineno[1]{%
  \patchAmsMathEnvironmentForLineno{#1}%
  \patchAmsMathEnvironmentForLineno{#1*}}%
\AtBeginDocument{%
  \patchBothAmsMathEnvironmentsForLineno{equation}%
  \patchBothAmsMathEnvironmentsForLineno{align}%
  \patchBothAmsMathEnvironmentsForLineno{flalign}%
  \patchBothAmsMathEnvironmentsForLineno{alignat}%
  \patchBothAmsMathEnvironmentsForLineno{gather}%
  \patchBothAmsMathEnvironmentsForLineno{multline}%
}

\title{Nonconvex Regularization for Feature Selection\\ in Reinforcement Learning \vspace{-10pt}}
\name{Kyohei Suzuki \qquad Konstantinos Slavakis\thanks{This work was supported by the Grants-in-Aid for Scientific
    Research (KAKENHI) under Grant Number 25K24422.} \vspace{-10pt} }

\address{ Institute of Science Tokyo, Japan\\ \small Department of Information and Communications Engineering\\[-3pt]
  \small Emails: \texttt{suzuki.k.439f@m.isct.ac.jp, slavakis@ict.eng.isct.ac.jp} \vspace{-10pt} }
\begin{document}
\ninept
\renewcommand{\baselinestretch}{1.0}
\setlength{\abovedisplayskip}{6pt}
\setlength{\belowdisplayskip}{5pt}
\setlength{\abovedisplayshortskip}{3pt}
\setlength{\belowdisplayshortskip}{4pt}

\setlength{\textfloatsep}{12pt}
\setlength{\intextsep}{12pt}

\setlength{\parskip}{0pt}
\setlist{nosep}

\maketitle
\sloppy

\begin{abstract}
  This work proposes an efficient batch algorithm for feature selection in reinforcement learning (RL) with theoretical
  convergence guarantees. To mitigate the estimation bias inherent in conventional regularization schemes, the first
  contribution extends policy evaluation within the classical least-squares temporal-difference (LSTD) framework by
  formulating a Bellman-residual objective regularized with the sparsity-inducing, nonconvex projected minimax
  concave (PMC) penalty. Owing to the weak convexity of the PMC penalty, this formulation can be interpreted as a
  special instance of a general nonmonotone-inclusion problem. The second contribution establishes novel convergence
  conditions for the forward-reflected-backward splitting (FRBS) algorithm to solve this class of problems. Numerical
  experiments on benchmark datasets demonstrate that the proposed approach substantially outperforms state-of-the-art
  feature-selection methods, particularly in scenarios with many noisy features.
\end{abstract}


\section{Introduction}\label{sec:intro}

Reinforcement learning (RL) plays a central role in contemporary machine learning, signal processing, and control
theory~\cite{sutton1998reinforcement, bertsekas2019reinforcement, arulkumaran2017deep}. The primary objective of RL is
for an agent to learn an optimal policy to control a system by minimizing a long-term loss, represented by the
Q-function. This learning occurs through interactions with the environment, which is typically modeled as a Markov
decision process (MDP). In most high-dimensional, real-world problems, explicitly representing the Q-function for all
possible states and actions is impractical due to the ``curse of dimensionality.'' A common solution is to approximate
the Q-function using a parametric (functional) representation. This, however, introduces a fundamental trade-off between
approximation accuracy and computational complexity: reducing the approximation error generally requires a large number
of features in the parametric model, which in turn increases computational demands.

Feature selection, achieved via a sparse representation over a large basis of functions, is an effective way to
alleviate this trade-off, mitigate overfitting, and improve sample efficiency. Most existing work on feature selection
in RL has focused on $\ell_1$-norm regularization to induce sparsity, including batch algorithms such as least-angle
regression (LARS)-TD~\cite{kolter2009regularization}, $\ell_1$-projected Bellman residual~(BR)~\cite{geist2011ell1}, and
basis-pursuit denoising (BPDN)~\cite{qin2014sparse} (see \cite{liu2015feature} and references therein), as well as
online algorithms~\cite{mahadevan2012sparse, song2018sparse, song2021online}.

The least absolute shrinkage and selection operator (LASSO)~\cite{tibshirani1996regression} is widely used for sparse
regression but suffers from estimation bias due to underestimating large-amplitude components~\cite{fan2001variable,
  osher2016sparse}. To mitigate this limitation, nonconvex penalties have been proposed~\cite{zhang2010nearly,
  selesnick2017sparse, chartrand2007exact}, with particular focus on those that preserve the overall convexity of the
cost function when combined with a least-squares model~\cite{selesnick2017sparse, bayram2016convergence,
  abe2020linearly, yukawa2023linearly}. In underdetermined settings, maintaining convexity requires the regularizer to
be nonseparable~\cite{selesnick2017sparse}. The projected minimax concave (PMC) penalty~\cite{yukawa2023linearly} meets
this requirement while allowing efficient implementation: the penalty itself is nonseparable, yet the internal objective
defining its Moreau envelope is additively separable if the seed function is separable (see Section
\ref{subsec:Projective Minimax Concave Penalty}). Despite these advantages, nonconvex sparsity-inducing penalties remain
unexplored in RL, even for batch algorithms. This raises a natural question: can such penalties be effectively employed
for feature selection in RL while ensuring convergence guarantees?

This paper provides an affirmative answer to the previous research question through a two-fold contribution. First, on
the algorithmic side, it introduces an efficient batch method for feature selection in RL by formulating an optimization
problem that augments the classical least-squares temporal-difference (LSTD) loss~\cite{bradtke1996linear,
  lagoudakis2003least} with the nonconvex PMC penalty. By exploiting the weak convexity of PMC, the problem is
reformulated into a tractable form and solved using the forward-reflected-backward splitting (FRBS)
method~\cite{malitsky2020forward}. Second, on the theoretical side, since the reformulated problem constitutes a
nonmonotone-inclusion task for which the convergence conditions in~\cite{malitsky2020forward} are not directly
applicable, extended conditions are developed to guarantee convergence for a broad class of nonmonotone-inclusion
problems, encompassing the proposed nonconvexly regularized LSTD setting as a special case. Numerical tests further
validate the approach, demonstrating that the proposed method substantially outperforms state-of-the-art batch methods,
even in scenarios with numerous noisy features.
\vspace{-1ex}

\section{Preliminaries}

Throughout the paper, let $\mathbb{R}$, $\mathbb{R}_{+}$, $\mathbb{R}_{++}$, and $\mathbb{N}$ denote the sets of real
numbers, nonnegative real numbers, strictly positive real numbers, and nonnegative integers, respectively. Let also
$\mathbb{N}_* \coloneqq \mathbb{N} \setminus \{0\}$. For $n_1, n_2\in \mathbb{N}_*$, with $n_1 \leq n_2$, define
$\overline{n_1, n_2} \coloneqq \{ n_1, n_1+1, \ldots, n_2\}$. Vector/matrix transposition is denoted by
$(\cdot)^{\top}$.  For any $\bm{x} \in \mathbb{R}^n$, the $\ell_p$ norm is defined by $\| \bm{x} \|_p \coloneqq \left(
\sum_{i=1}^n \left| x_i \right|^p \right)^{1 / p}$ for $p \ge 1$, and the $\ell_{\infty}$ norm is defined by
$\|\bm{x}\|_{\infty} \coloneqq \max \{|x_1|, |x_2|, \ldots, |x_n|\}$.  Let also $\bm{O}_{m \times n}$ and $\bm{0}_n$
denote the $m \times n$ zero matrix and $n$-dimensional zero vector, respectively.
\vspace{-1ex}
\vspace{-1ex}

\subsection{Background on reinforcement learning}
\label{subsec:Background on reinforcement learning}

In RL problems, the interaction between an environment and an agent is modeled as an MDP defined as the tuple
$(\mathfrak{S}, \mathfrak{A}, \mathbb{P}, g, \gamma )$, where, for simplicity, $\mathfrak{S} \subset \mathbb{R}^D$ is a
finite state space for some $D \in \mathbb{N}_*$, $\mathfrak{A}$ is a finite action space, and $\mathbb{P}$ is a
Markovian transition model~\cite{lagoudakis2003least}. Here, $\mathbb{P}(\bm{s}^{\prime} \mid \bm{s}, a)$ is
the probability of transitioning to state $\bm{s}^{\prime}$ when taking action $a$ in state $\bm{s}$. Moreover, $g
\colon \mathfrak{Z} \rightarrow \mathbb{R}_+$ is the one-step loss, and $\gamma \in (0, 1)$ is the discount factor. A
deterministic stationary policy $\pi \colon \mathfrak{S} \rightarrow \mathfrak{A}$ specifies the action $\pi(\bm{s})$
that the agent takes in state $\bm{s}$.

Following~\cite{lagoudakis2003least}, the expected long-term loss is defined as $Q^{\pi}\colon \mathfrak{Z} \to
\mathbb{R}_+ \colon (\bm{s}, a) \mapsto \mathbb{E} \{ \sum_{t = 0}^{+\infty} \gamma^{t} g (\bm{s}_t, \bm{a}_t) \mid
\bm{s}_0 = \bm{s}, a_0 = a \}$, where the expectation is taken with respect to (w.r.t.) the trajectories generated by
$\mathbb{P}$ and $\pi$, and $(\bm{s}_t, a_t)$ denotes the state-action random variables at time $t$. It is well known to
satisfy the following (fixed-point) Bellman equation: $\forall (\bm{s}, a) \in \mathfrak{S} \times \mathfrak{A}$,
\begin{alignat}{2}
  Q^{\pi} (\bm{s}, a)
  & = && ( T_{\pi} Q^{\pi} ) (\bm{s}, a) \notag \\
  & {} \coloneqq {} && g(\bm{s}, a) + \gamma \mathbb{E}_{ \bm{s}^{\prime} \sim \mathbb{P}(\cdot \mid \bm{s},
      a)} \{ Q^{\pi}(\bm{s}^{\prime}, \pi(\bm{s}^{\prime}))\}, \label{eq:Bellman_equation}
\end{alignat}
where $\mathbb{E}_{ \bm{s}^{\prime} \sim \mathbb{P}(\cdot \mid \bm{s}, a)} \{\cdot\}$
denotes the conditional expectation over all potential subsequent states $\bm{s}'$ whenever the RL agent takes action
$a$ while in state $\bm{s}$.
\vspace{-1ex}
\subsection{Least-squares temporal difference (LSTD)}

When the number of states is exceedingly large or the state space is continuous, evaluating $Q^{\pi}$ becomes
computationally prohibitive. A common remedy is to approximate $Q^{\pi}$ by a ``linear function,'' $Q^{\pi}(\bm{s}, a)
\approx \bm{w}^{\top} \bm{\phi} (\bm{s}, a)$, which is linear in $\bm{w} \in \mathbb{R}^n$---the vector collecting all
learnable parameters of the approximation---for some $n \in \mathbb{N}_*$, where $\bm{\phi}(\bm{s}, a) \in \mathbb{R}^n$
denotes the feature vector at $(\bm{s}, a) \in \mathfrak{Z}$.
The Bellman-residual family of algorithms solves for a $\bm{w}$ such that (s.t.) its corresponding Q-function satisfies \eqref{eq:Bellman_equation}
approximately:
  {\jot=1pt
    \begin{align}
      \!\!\mathrm{find}\ & \bm{w} \in \mathbb{R}^n \notag\\
      \!\!\text{s.t.}\ & \bm{w} \in \argmin_{\bm{u} \in \mathbb{R}^n}
      \frac{1}{2} \sum_{(\bm{s}, a) \in \mathfrak{Z}} \!\left[\bm{u}^{\top} \bm{\phi}(\bm{s}, a) - ( T_{\pi}\, \bm{w}^{\top}
        \bm{\phi} ) (\bm{s}, a) \right]^2 \,\!\!.\!\!
      \label{eq:Bellman_residual}
    \end{align}
  }
Since $\mathbb{P}$ is in general unavailable,
LSTD~\cite{bradtke1996linear,lagoudakis2003least} approximates \eqref{eq:Bellman_residual} via $m$ samples $(s_i, r_i, s_i')_{i= 1}^m$:
\begin{align}
  \mathrm{find}~\bm{w} \in \mathbb{R}^n\ \text{s.t.}\ \bm{w} \in \argmin_{\bm{u} \in \mathbb{R}^n} \frac{1}{2}
  \| \widetilde{\bm{\Phi}} \bm{u} - (\widetilde{\bm{g}} + \gamma \widetilde{\bm{\Phi}}' \bm{w} )
  \|_2^2, \label{eq:LSTD_formulation}
\end{align}
where the $m\times n$ matrices $\widetilde{\bm{\Phi}} \coloneqq [\bm{\phi}(s_1, a_1), \ldots, \bm{\phi}(s_m,
  a_m)]^{\top}$, $\widetilde{\bm{\Phi}}' \coloneqq [ \bm{\phi}(s'_1, \pi(s'_1)), \ldots, \bm{\phi}(s'_m, \pi(s'_m))
]^{\top}$, and the $m\times 1$ vector $\widetilde{\bm{g}} \coloneqq [ g(s_1, a_1), \ldots, g(s_m, a_m) ]^{\top}$. A
solution to \eqref{eq:LSTD_formulation} can be obtained analytically by $\bm{w}_* = \widetilde{\bm{A}}^{-1}
\widetilde{\bm{b}}$, where
\begin{align}
  \widetilde{\bm{A}} \coloneqq \widetilde{\bm{\Phi}}^{\top} (\widetilde{\bm{\Phi}} - \gamma
  \widetilde{\bm{\Phi}}'), \quad \widetilde{\bm{b}} \coloneqq \widetilde{\bm{\Phi}}^{\top}
  \widetilde{\bm{g}}. \label{def:A_tilde_b_tilde}
\end{align}
If $\widetilde{\bm{A}}$ is not invertible, one can instead employ the Moore-Penrose pseudoinverse of
$\widetilde{\bm{A}}$ or ridge regression~\cite{lagoudakis2003least, yu2009convergence}.
\vspace{-1ex}

\subsection{Mathematical preliminaries}\label{subsec:selected_elements}

A function $f\colon \mathbb{R}^n \to (-\infty, +\infty] \coloneqq \mathbb{R} \cup \{+ \infty\}$ is proper if $\dom f
  \coloneqq \{\bm{x} \in \mathbb{R}^n \mid f(\bm{x}) < +\infty\} \neq \emptyset$.  A function $f\colon \mathbb{R}^n \to
  (-\infty, +\infty]$ is convex if $f(a \bm{x} + (1-a) \bm{\xi}) \leq a f(\bm{x})+(1-a) f(\bm{\xi})$ for any $\bm{x},
    \bm{\xi} \in \mathbb{R}^n$ and any $a \in(0,1)$.  Given a proper function $f \colon \mathbb{R}^n \to (-\infty,
    +\infty]$, the single-valued proximity operator of $f$ of index $\tau > 0$ is defined as $\sProx_{\tau f}\colon
      \mathbb{R}^n \to \mathbb{R}^n \colon \bm{x} \mapsto \argmin_{\bm{\xi} \in \mathbb{R}^n} f(\bm{\xi}) + \|\bm{x} -
      \bm{\xi}\|_2^2 / (2 \tau)$~\cite{yukawa2025monotone}, whenever $f + \|\bm{x} - \cdot\|_2^2 / (2\tau)$ has a unique
      minimizer for every fixed $\bm{x} \in \mathbb{R}^n$.\footnote{If $f$ is nonconvex, the proximity operator is often
      defined as a set-valued operator. This paper focuses on the case when the proximity operator is a single-valued
      operator, and the notation $\sProx$ is used to emphasize that.}  Given a proper function $f\colon \mathbb{R}^n \to
      (-\infty, +\infty]$, the Moreau envelope of $f$ of index $\tau > 0$ is defined as ${}^{\tau} \! f \colon
        \mathbb{R}^n \to \mathbb{R} \colon \bm{x} \mapsto \min_{\bm{\xi} \in \mathbb{R}^n} f(\bm{\xi}) + \|\bm{x} -
        \bm{\xi}\|_2^2 / (2 \tau)$~\cite{moreau1962decomposition, moreau1962fonctions, bauschke2017convex}. For any
        nonempty closed convex set $C \subset \mathbb{R}^n$, $P_C$ denotes the (metric) projection operator onto $C$.

A set-valued mapping $T\colon \mathbb{R}^n \to 2^{\mathbb{R}^n}$ is called monotone if $\langle \bm{x} - \bm{y}, \bm{u}
- \bm{v} \rangle_2 \ge 0$, $\forall (\bm{x}, \bm{u}), (\bm{y}, \bm{v}) \in \gra T$, where $\gra T \coloneqq \{(\bm{x},
\bm{u}) \in \mathbb{R}^n \times \mathbb{R}^n \mid \bm{u} \in T(\bm{x})\}$ denotes the graph of $T$.  A monotone operator
$T\colon \mathbb{R}^n \to 2^{\mathbb{R}^n}$ is maximally monotone if no other monotone operator has its graph containing
$\gra T$.  If $T - \rho \Id$ is (maximally) monotone for some $\rho \in \mathbb{R}$, $T$ is (maximally) $\rho$-monotone
\cite{bauschke2021generalized}, also known as $\rho$-strongly monotone when $\rho > 0$ and as $|\rho|$-hypomonotone when
$\rho < 0$.
\vspace{-1ex}

\subsection{Projective minimax concave penalty}\label{subsec:Projective Minimax Concave Penalty}

The MC penalty \cite{zhang2010nearly} with index $\tau > 0$ is defined as $\Psi_\tau^{\mathrm{MC}}\colon \mathbb{R}^n
\to [0,+\infty) \colon \bm{x} \mapsto \sum_{i=1}^n \psi_\tau^{\mathrm{MC}} \left(x_i\right)$, where
  $\psi_\tau^{\mathrm{MC}} \colon \mathbb{R} \to [0,+\infty) \colon x \mapsto
    \begin{cases}
        |x|- x^2 / (2 \tau), & \text { if }|x| \leq \tau, \\
        \tau / 2, & \text { if }|x|>\tau. \label{def:MC}
    \end{cases}$.
  The MC penalty can be expressed as the Moreau-enhanced model of the $\ell_1$ norm \cite{abe2020linearly},
  \textit{i.e.}, the difference between the $\ell_1$ norm and its Moreau envelope \cite{selesnick2017sparse}:
  \begin{equation}
    \Psi_{\tau}^{\mathrm{MC}} = \|\cdot\|_1 - \hspace{.3em}^{\tau} \|\cdot\|_1, \label{eq:MC_L1_Moreau}
  \end{equation}
  The MC penalty is known to be weakly convex, \textit{i.e.}, it becomes convex by adding the squared $\ell_2$ norm
  multiplied by some positive constant \cite{selesnick2017sparse}. For some linear subspace $\mathcal{M} \subset
  \mathbb{R}^n$, the projective minimax concave (PMC) penalty~\cite{yukawa2023linearly} is defined as\footnote{While the
  PMC penalty is originally proposed for $\mathcal{M} \coloneqq \mathrm{null}^{\perp} \bm{M}$, this paper considers the case in which $\mathcal{M} $ is a general subspace in $\mathbb{R}^n$.}
    \begin{equation}
    \Psi_{\tau, \mathcal{M}}^{\mathrm{PMC}} \coloneqq \|\cdot\|_1 - \hspace{.3em}^{\tau} \|\cdot\|_1 (P_{\mathcal{M}}
    \cdot ). \label{def:PMC}
  \end{equation}
    While the PMC penalty is additively nonseparable due to the presence of $P_{\mathcal{M}}$, the second term of
    \eqref{def:PMC} can be rewritten as: $\bm{x} \mapsto \|\cdot\|_1 (P_{\mathcal{M}} \bm{x}) = \min_{\bm{u} \in
      \mathbb{R}^n} [\|\bm{u}\|_1 + (1 / (2 \tau)) \|P_{\mathcal{M}} \bm{x} - \bm{u}\|_2^2]$, where the internal
    objective function is additively separable as a function of $\bm{u}$. This structure enables the PMC penalty to
    simultaneously preserve overall convexity and support efficient implementation without the introduction of
    additional variables (see \cite{yukawa2023linearly}).

  The PMC penalty coincides with the MC penalty on $\mathcal{M}$ and with the $\ell_1$ norm on
  $\mathcal{M}^{\perp}$~\cite[Proposition 1]{yukawa2023linearly}, and hence it restricts the debiasing effect of MC to
  $\mathcal{M}$. Owing to this, the convexity of the LS loss regularized by the PMC penalty is guaranteed even in the
  underdetermined case, unlike the case of MC penalty, as shown by the following fact.

  \begin{fact}{\cite[Proposition 2]{yukawa2023linearly}}
    \label{fact:PMC_convexity}
    Let $\bm{M} \in \mathbb{R}^{p \times q}$, $\bm{y} \in \mathbb{R}^p$, and $\mathcal{M} \coloneqq
    \mathrm{null}^{\perp} \bm{M} (= \range \bm{M}^{\top})$, for some $p, q \in \mathbb{N}_*$.  Then, function $f\colon
    \mathbb{R}^q \to \mathbb{R}_+ \colon \bm{x} \mapsto (1/2) \|\bm{M} \bm{x} - \bm{y}\|_2^2 - \mu \hspace{.3ex}^{\tau}
    \|\cdot\|_1 (P_{\mathcal{M}}\bm{x})$ is convex if and only if $\mu \tau^{-1} \le \lambda_{\min}^{++} (\bm{M}^{\top}
    \bm{M})$, where $\lambda_{\min}^{++} (\cdot)$ denotes the smallest strictly-positive eigenvalue.
  \end{fact}
\vspace{-1ex}
\vspace{-1ex}

\section{Main Results}

The first contribution incorporates the PMC penalty as a regularizer into the LSTD formulation, reformulating the
resulting problem as finding the zeros of the sum of a monotone Lipschitz operator and a hypomonotone operator. An
FRBS-based optimization algorithm is proposed to solve this problem. The second contribution establishes convergence
conditions for the algorithm in a general nonmonotone-inclusion setting. Theoretical results are stated here without
proof, with full proofs to be provided elsewhere.

\subsection{Problem formulation}

To identify a sparse weight vector $\bm{w}$, this paper proposes to combine the LSTD formulation in
\eqref{eq:LSTD_formulation} with the PMC penalty in \eqref{def:PMC}, yielding:
  {\jot=1pt
  \begin{align}
    \mathrm{find}\ {} & {} \bm{w} \in \mathbb{R}^n \notag \\
    \mathrm{s.t.}\ {} & {} \bm{w} \in \argmin_{\bm{u} \in \mathbb{R}^n} \frac{1}{2} \| \widetilde{\bm{\Phi}} \bm{u} -
    (\widetilde{\bm{g}} + \gamma \widetilde{\bm{\Phi}}' \bm{w})\|_2^2 + \mu \Psi_{\tau,
      \mathcal{M}}^{\mathrm{PMC}}(\bm{u}), \label{eq:optimization_problem_original}
  \end{align}
  }
where $\mu, \tau \in \mathbb{R}_{++}$ and $\mathcal{M}$ is a linear subspace of $\mathbb{R}^n$.  When $\tau \rightarrow
+ \infty$, the PMC penalty reduces to the $\ell_1$ norm, the objective function of the ``lower-level'' problem
w.r.t.\ $\bm{u}$ becomes convex, and \eqref{eq:optimization_problem_original} takes the form of
LARS-TD~\cite{kolter2009regularization}. However, even in this setting, where the ``lower-level'' minimization task is
convex, it is known that the overall problem \eqref{eq:optimization_problem_original} w.r.t.\ $\bm{w}$ cannot be
cast as a convex minimization one~\cite{kolter2009regularization}.

Nevertheless, the objective function of the ``lower-level'' problem in \eqref{eq:optimization_problem_original}
w.r.t.\ $\bm{u}$ can be made convex, for any fixed $\bm{w}$, by a suitable choice of $\mu$ and $\tau$. By
Fact~\ref{fact:PMC_convexity}, this convexity is preserved even when $\widetilde{\bm{\Phi}}^{\top}
\widetilde{\bm{\Phi}}$ is singular, unlike the case with the MC penalty. Exploiting this induced convexity of the
``lower-level'' problem, \eqref{eq:optimization_problem_original} can be reformulated as a nonmonotone inclusion
problem, formalized in the following proposition.

\begin{proposition}\label{PROP:REFORMULATION}
  Let $\widetilde{\bm{\Phi}}^{\top} \widetilde{\bm{\Phi}} = \bm{V} \bm{\Lambda} \bm{V}^{\top}$ be the eigenvalue
  decomposition of $\widetilde{\bm{\Phi}}^{\top} \widetilde{\bm{\Phi}}$, where $\bm{V} \in \mathbb{R}^{n \times n}$ is
  an orthogonal matrix, and $\bm{\Lambda} = \diag (l_1, l_2, \ldots, l_n)$, with $l_1 \ge l_2 \ge \ldots \ge l_n \ge 0$.
  For some $q \in \overline{1, n}$, let $\bm{V} = [ \bm{V}_{\overline{1,q}}, \bm{V}_{\overline{q+1, n}}]$, where
  $\bm{V}_{\overline{1,q}} \in \mathbb{R}^{n \times q}$ and $\bm{V}_{\overline{q+1, n}} \in \mathbb{R}^{n \times (n -
    q)}$, and $\bm{\Lambda} = \left[ \begin{smallmatrix} \bm{\Lambda}_{\overline{1,q}} & \bm{O}_{q \times q}
      \\ \bm{O}_{(n-q) \times (n - q)} & \bm{\Lambda}_{\overline{q+1,n}} \end{smallmatrix} \right]$, where
  $\bm{\Lambda}_{\overline{1,q}} \in \mathbb{R}^{q \times q}$ and $\bm{\Lambda}_{\overline{q+1,n}} \in \mathbb{R}^{(n -
    q) \times (n - q)}$. Assume that $\mathcal{M} \coloneqq \mathrm{span} (\bm{V}_{\overline{1,q}})$ and
  \begin{equation}
    \mu \tau^{-1} \le \max \{l_q, \lambda_{\min}^{++} (\widetilde{\bm{\Phi}}^{\top} \widetilde{\bm{\Phi}})
    \}. \label{eq:tau_condition_1}
  \end{equation}
  Then, $\bm{w} \in \mathbb{R}^n$ is a solution, if exists, of \eqref{eq:optimization_problem_original} if and
  only if
  \begin{align}
    \bm{0}_N \in {} & {} T(\bm{w}) + \mu \partial \| \cdot \|_1 (\bm{w})
    \,, \label{eq:optimization_problem_reformulated}
  \end{align}
  where $T \colon \mathbb{R}^n \rightarrow \mathbb{R}^n \colon \bm{w} \mapsto T(\bm{w})$ with
  \begin{align}
    T(\bm{w}) \coloneqq \widetilde{\bm{A}} \bm{w} - \widetilde{\bm{b}} - \mu \tau^{-1} P_{\mathcal{M}} (\Id -
    \sProx_{\tau \|\cdot\|_1} )(P_{\mathcal{M}} \bm{w}) \,. \label{map.T}
  \end{align}
\end{proposition}


\begin{remark}
  As discussed in Section~\ref{subsec:Projective Minimax Concave Penalty}, the debiasingeffect of the PMC penalty is
  confined to the subspace $\mathcal{M}$. To maximize this debiasing effect, $\mathcal{M}$ should have the largest
  possible dimension, which is achieved by setting $\mathcal{M} \coloneqq \bm{V}_{\overline{1,q}} =
  \mathrm{null}^{\perp} \widetilde{\bm{\Phi}}$, with $q$ satisfying $l_q = \lambda_{\min}^{++}
  (\widetilde{\bm{\Phi}}^{\top} \widetilde{\bm{\Phi}})$. However, if $\lambda_{\min}^{++} (\widetilde{\bm{\Phi}}^{\top}
  \widetilde{\bm{\Phi}})$ is very small, $\tau$ must be sufficiently large to meet \eqref{eq:tau_condition_1}, which can
  degrade performance since the PMC penalty approaches the $\ell_1$ norm for large $\tau$. Therefore, in practice,
  choosing an appropriate $q$, even when $l_q < \lambda_{\min}^{++} (\widetilde{\bm{\Phi}}^{\top}
  \widetilde{\bm{\Phi}})$, may yield better performance than fully setting $\mathcal{M} = \mathrm{null}^{\perp}
  \widetilde{\bm{\Phi}}$.
\end{remark}
\vspace{-1ex}
\vspace{-1ex}

\subsection{Recasting and solving \eqref{eq:optimization_problem_reformulated}}\label{sec:recast.problem}

Since the form of $T$ in \eqref{map.T} renders $T + \mu \partial \|\cdot\|_1$ nonmonotone, the popular forward-backward
splitting~\cite{mercier1979lectures, combettes2004solving} and Douglas-Rachford~\cite{lions1979splitting,
  eckstein1992douglas, combettes2004solving} methods cannot guarantee convergence to a solution of
\eqref{eq:optimization_problem_reformulated}. To surmount this obstacle, \eqref{eq:optimization_problem_reformulated} is
recast and solved using the FRBS method~\cite{malitsky2020forward}. While the convergence guarantees of FRBS are
established in~\cite{malitsky2020forward} only for the case where $T + \mu \partial \|\cdot\|_1$ is monotone, these
conditions are extended here to accommodate the nonmonotone case (see Section~\ref{subsec: Generalized Case}).

To this end, \eqref{eq:optimization_problem_reformulated} is recast as
\begin{align}
  \bm{0}_N \in (\alpha T + \Id) (\bm{w}) + (\alpha \mu \partial \|\cdot\|_1 - \Id)
  (\bm{w}) \,, \label{eq:reforemulated_problem}
\end{align}
where $\alpha \in (0, (\| \widetilde{\bm{A}} \|_2 + \mu \tau^{-1} )^{-1} ]$. Then, it can be shown that $\alpha T + \Id$
  is $\beta$-Lipschitz continuous and maximally monotone for $\beta \coloneqq \alpha (\| \widetilde{\bm{A}} \|_2 + \mu
  \tau^{-1}) + 1$.
   Moreover,
  $\alpha \mu \partial \|\cdot\|_1 - \Id$ is maximally $(-1)$-monotone since $(\alpha \mu \partial \|\cdot\|_1 - \Id) -
  (-1)\Id = \alpha \mu \partial \|\cdot\|_1$ is maximally monotone.

\begin{algorithm}[t!]
\caption{The FRBS method for solving \eqref{eq:optimization_problem_original}} \label{alg:forward_reflected_backward}

\begin{algorithmic}[1]

      \Require $(\widetilde{\bm{A}}, \widetilde{\bm{b}})$ from \eqref{def:A_tilde_b_tilde}, $\mu \in \mathbb{R}_{++}$,
      $q \in \overline{1, n}$ (see Proposition~\ref{PROP:REFORMULATION})

      \State Set $\tau$, $\alpha$, and $(\eta_k)_{k \in \mathbb{N}}$ by Proposition~\ref{PROP:CONVERGENCE}

      \State Initialize $\bm{w}_{-1}, \bm{w}_0 \in \mathbb{R}^n$

      \State Set $\bm{u}_{-1} \coloneqq \alpha (\widetilde{\bm{A}} \bm{w}_{-1} - \widetilde{\bm{b}} - \mu \tau^{-1}
      (\bm{w}_{-1} - \Soft_{\tau}(\bm{w}_{-1}))) + \bm{w}_{-1}$ \For{$k = 0, 1, 2, \dots$} \State $\bm{u}_k \coloneqq
      \alpha (\widetilde{\bm{A}} \bm{w}_k - \widetilde{\bm{b}} - \mu \tau^{-1} (\bm{w}_k - \Soft_{\tau}(\bm{w}_k))) +
      \bm{w}_{k}$

      \State $\bm{w}_{k+1} \coloneqq \Soft_{\alpha \mu / (1 - \eta_k)} ((1 - \eta_k)^{-1}( \bm{w}_k - 2\eta_k \bm{u}_k +
      \eta_k \bm{u}_{k-1} ))$

      \EndFor

    \end{algorithmic}

\end{algorithm}

\begin{algorithm}[t!]
    \caption{Approximate policy iteration}\label{alg:policy_iteration_PMC}

    \begin{algorithmic}[1]

      \Require $\pi_0$, $\gamma \in [0, 1]$

      \For{$k = 0, 1, 2, \dots$}

      \State \textbf{Collect data:} Interact with the environment using $\pi_k$ to generate samples $(s_i, r_i, s_i')_{i
      = 1}^m$

      \State Define $(\widetilde{\bm{A}}, \widetilde{\bm{b}})$ by \eqref{def:A_tilde_b_tilde}

      \State \textbf{Policy Evaluation:}

      \State \hspace{1em} Obtain $\hat{\bm{w}}$ by Algorithm~\ref{alg:forward_reflected_backward}

      \State \hspace{1em} Set $\hat{Q}^{\pi_k} \coloneqq \tilde{\bm{\Phi}} \hat{\bm{w}}$

      \State \textbf{Policy Improvement:}

      \State \hspace{1em} $\pi_{k+1}(\bm{s}) \coloneqq \argmin_{\bm{a} \in \mathfrak{A}} \hat{Q}^{\pi_k}(\bm{s}, a) , ~ \forall \bm{s}
      \in \mathfrak{S}$

      \EndFor

    \end{algorithmic}

\end{algorithm}

Applying the FRBS method to \eqref{eq:reforemulated_problem} yields Algorithm~\ref{alg:forward_reflected_backward}.
Here, for any $\tau > 0$, the soft-shrinkage operator is defined as
$\Soft_{\tau} \coloneqq \sProx_{\tau\| \cdot \|_1} \colon \mathbb{R}^n \rightarrow \mathbb{R}^n \colon [x_1, \ldots,
  x_n]^{\top} \mapsto [\soft_{\tau}(x_1), \ldots, \soft_{\tau}(x_n)]^{\top}$, with $\soft_{\tau}(x) \coloneqq
\sign(x) \max \{ |x|-\tau, 0 \}$ and $\sign(x) \coloneqq 1$, if $x \ge 0$, while $\sign(x) \coloneqq -1$, if $x < 0$,
$\forall x\in \mathbb{R}$. The following proposition provides convergence guarantees for
Algorithm~\ref{alg:forward_reflected_backward}.

\begin{proposition}[Convergence of Algorithm \ref{alg:forward_reflected_backward}] \label{PROP:CONVERGENCE}
  Let $\beta \coloneqq \alpha (\|\widetilde{\bm{A}}\|_2 $ $+ \mu \tau^{-1}) + 1$. Assume the following:
  \begin{enumerate}[wide=0pt]
    \renewcommand{\theenumi}{\normalfont{(C-\arabic{enumi})}}
    \renewcommand{\labelenumi}{\normalfont{(C-\arabic{enumi})}}
  \item \label{cond:C-1} $\mu$, $\tau$, and $q$ satisfy \eqref{eq:tau_condition_1}.
  \item \label{cond:C-3} $\alpha \in (0, (\|\widetilde{\bm{A}}\|_2 + \mu \tau^{-1})^{-1}]$.
\item \label{cond:C-2} $(\eta_k)_{k \in \mathbb{N}} \subseteq [\epsilon, (1 -2 \epsilon) / (2 \beta)]$ for some
  $\epsilon \in (0, 1 / (2 (\beta + 1))]$.
  \end{enumerate}
  Then, for arbitrarily chosen $\bm{w}_{-1}, \bm{w}_0 \in \mathbb{R}^n$, the sequence $(\bm{w}_k)_{k \in \mathbb{N}}$
  produced by Algorithm~\ref{alg:forward_reflected_backward} converges to a solution, if exists, of
  \eqref{eq:optimization_problem_original}.
  \end{proposition}


An approximate policy-iteration (PI) method, based on Algorithm~\ref{alg:forward_reflected_backward}, is presented in
Algorithm~\ref{alg:policy_iteration_PMC}. Along the lines of~\cite[Theorem 3.1]{lagoudakis2003least}, the following
proposition shows that the sequence $(\pi_{k})_{k \in \mathbb{N}}$ generated by Algorithm~\ref{alg:policy_iteration_PMC}
either converges or oscillates within a region of the policy space, where the suboptimality of the resulting policies is
bounded by the approximation error of Algorithm~\ref{alg:forward_reflected_backward}.


\begin{proposition}
  \label{prop:convergence_API}
  Let $(\pi_k)_{k \in \mathbb{N}}$ and $(\hat{Q}^{\pi_k})_{k \in \mathbb{N}}$ be sequences generated by Algorithm
  \ref{alg:policy_iteration_PMC}.  Suppose that there exist $\delta_1, \delta_2 \in \mathbb{R}_{++}$ s.t.\ $\|
  \hat{Q}^{\pi_k} - Q^{\pi_k} \|_{\infty} \le \delta_1$ and $\| T_{\pi_{k+1}} \hat{Q}^{\pi_k} - T_*
  \hat{Q}^{\pi_k}\|_{\infty} \le \delta_2$, $\forall k \in \mathbb{N}_*$, where $(T_* \hat{Q}^{\pi_k})(\bm{s}, a) \coloneqq
  g(\bm{s}, a)+\gamma \mathbb{E}_{\bm{s}^{\prime} \sim \mathbb{P}(\cdot \mid \bm{s}, a)}\{\min_{a^{\prime}}
  \hat{Q}^{\pi_k}(\bm{s}^{\prime}, a^{\prime})\},~ \forall (\bm{s}, a) \in \mathfrak{Z}$.
  Here, $Q^{\pi_k}$ and $T_{\pi_{k+1}}$ are defined in the same way as in Section \ref{subsec:Background on reinforcement learning}.
  Then,
  {\abovedisplayskip=2pt
  \belowdisplayskip=2pt
  \abovedisplayshortskip=1pt
  \belowdisplayshortskip=1pt
  \begin{align}
    \limsup_{k \rightarrow + \infty} \| \hat{Q}^{\pi_k} - Q^{*} \|_{\infty} \le \frac{2 \gamma \delta_1 +
      \delta_2}{(1 - \gamma)^2} \,,
  \end{align}
  }
  where $Q^{*}(\bm{s},a) \coloneqq \min_\pi Q^\pi(\bm{s},a),~ \forall (\bm{s}, a) \in \mathfrak{Z}$.
\end{proposition}

\subsection{Solving a general nonmonotone-inclusion problem} \label{subsec: Generalized Case}

For a finite-dimensional Hilbert space $(\mathcal{H}, \langle \cdot, \cdot \rangle_{\mathcal{H}})$, consider the problem
\begin{align}
  \mathrm{find}\ x \in \mathcal{H}\ \mathrm{s.t.}\ 0 \in (A + B) (x), \label{eq:general_problem}
\end{align}
where $A\colon \mathcal{H} \rightarrow 2^{\mathcal{H}}$ is maximally $\rho$-monotone for some $\rho \in \mathbb{R}$, and
$B\colon \mathcal{H} \rightarrow \mathcal{H}$ is monotone and $L_{B}$-Lipschitz continuous. Problem
\eqref{eq:general_problem} encompasses \eqref{eq:reforemulated_problem}, as can be readily verified for the case where
$A = \alpha \mu \partial \|\cdot\|_1 - \Id$ with $\rho = -1$, and $B = \alpha T + \Id$ with $L_B = \beta$---see
Section~\ref{sec:recast.problem}.

To solve \eqref{eq:general_problem} apply the FRBS method: for the arbitrarily chosen initial points $x_{-1}, x_0 \in
\mathcal{H}$, generate $(x_k)_{k \in \mathbb{N}} \subset \mathcal{H}$ as: $\forall k \in \mathbb{N}$,
\begin{align}
  \!\!x_{k+1} \!\coloneqq \!J_{\eta_k A} [\, x_k - \eta_k B (x_k) - \eta_{k - 1} (\,\! B(x_k) - B (x_{k - 1})\,\! )\,\!
  ]\,, \label{eq:FRBS_algorithm_generalized}
\end{align}
where $J_{\eta_k A}$ denotes the resolvent of $\eta_k A$.\footnote{For any $A\colon \mathcal{H} \rightarrow
2^{\mathcal{H}}$, the resolvent of $A$ is defined as $J_A \coloneqq (\Id + A)^{-1}$, where $\Id$ is the identity
operator in $\mathcal{H}$.} Although the convergence of the FRBS method is established only for $\rho \ge
1$~\cite{malitsky2020forward}, the following theorem guarantees convergence even when $\rho$ is possibly negative.

\begin{theorem}
  \label{THEOREM:CONVERGENCE_GUARANTEE}
  Let $A\colon \mathcal{H} \rightarrow 2^{\mathcal{H}}$ be maximally $\rho$-monotone for some $\rho \in \mathbb{R}$,
  while $B\colon \mathcal{H} \rightarrow \mathcal{H}$ is monotone and $L_{B}$-Lipschitz continuous. Suppose that $(A +
  B)^{-1} (0) \neq \emptyset$, and let sequence $(\eta_k)_{k \in \mathbb{N}} \subseteq [\epsilon, (1 - 2 \epsilon) / (2
    L)] \cap (-1/\rho, +\infty)$, for some $\epsilon \in (0, 1 / ( 2(L_B + 1) ) ]$.  Then, for arbitrarily fixed initial
    points $x_{-1}, x_0 \in \mathcal{H}$, the sequence $(x_k)_{k \in \mathbb{N}} \subset \mathcal{H}$ generated by
    \eqref{eq:FRBS_algorithm_generalized} converges to a point in $(A + B)^{-1} (0)$.
\end{theorem}


Finally, the following proposition provides a closed-form expression of the resolvent of $\eta_k(\alpha \mu \partial
\|\cdot\|_1 - \Id)$.

\begin{proposition}
  \label{PROP:RESOLVENT_OF_A}
  Set $\mu$, $\alpha$, and $\eta_k$ so that assumptions in Proposition \ref{PROP:CONVERGENCE} are satisfied. Then,
  \begin{align}
    J_{\eta_k(\alpha \mu \partial \|\cdot\|_1 - \Id)} = \Soft_{\alpha \mu / (1 - \eta_k)} \circ\, (1 - \eta_k)^{-1} \Id
    \,.
  \end{align}
\end{proposition}


\section{Numerical Tests}

The performance of the proposed method is validated by a classical benchmark RL task called the $20$-state
MDP~\cite{lagoudakis2003least}.  Two actions, ``left'' and ``right'' are available, and the success probability of
either action is $0.9$.  When an action fails, the state changes to the opposite direction.  A reward is $1$ only at
states $1$ and $20$ and zero at anywhere else.  The optimal policy is to go left in states $1$--$10$ and right in states
$11$--$20$.
The value function for any policy $\pi$ is defined as $V^{\pi} \colon \mathfrak{S} \to \mathbb{R}_+\colon s \mapsto Q^{\pi}(s, \pi(s))$, $\forall s \in \mathfrak{S}$.
The optimal value function is defined as $V^* \colon \mathfrak{S} \to \mathbb{R}_+\colon s \mapsto \min_{\pi} V^{\pi}(s) =   \min_{a \in \mathfrak{A}} Q^*(s, a)$, $\forall s \in \mathfrak{S}$ \cite{bertsekas2019reinforcement}.
For this MDP, the optimal value function $V^*$ can be easily obtained analytically by solving \eqref{eq:Bellman_equation} because $\mathbb{P}$ is
available. Being scalar, state $s$ is not bold-faced.

For any $(s, a) \in \mathfrak{S} \times \mathfrak{A}$, the $n\times 1$ feature vector is constructed as:
$\begin{aligned}
  \phi (s,a) \coloneqq
  \begin{cases}
      [ \bm{\varphi}^{\top} (s), \bm{\varepsilon}^{\top}, \bm{0}_{n / 2}^{\top} ]^{\top}, & \text{if}\ a =
      \text{``left''}, \\
      [\bm{0}_{n / 2}^{\top}, \bm{\varphi}^{\top} (s), \bm{\varepsilon}^{\top} ]^{\top}, & \text{if}\ a =
      \text{``right''}, \\
  \end{cases}
\end{aligned}$ 
where $\bm{\varphi}(s) \coloneqq [1, \varphi_1^{\mathrm{RBF}}(s), \ldots,
  \varphi_{n_{\mathrm{RBF}}}^{\mathrm{RBF}}(s)]$, the Gaussian kernel $\varphi_i^{\mathrm{RBF}}(s) \coloneqq \exp (\, (s
- c_i)^2 / \varsigma\, )$, $i \in \overline{ 1, n_{\mathrm{RBF}} }$, with centers $\{ c_i \}_{i=1}^{n_{\mathrm{RBF}}}$
evenly aligned, and $\bm{\varepsilon} \in \mathbb{R}^{n_{\varepsilon}}$ is the realization of a vector-valued random
variable, representing irrelevant features, that follows the normal distribution $\mathcal{N}(\bm{0}_{n_{\varepsilon}},
0.1 \bm{I}_{n_{\varepsilon}})$.  Here, $n_{\mathrm{RBF}}, n_{\varepsilon} \in \mathbb{N}_{*}$ s.t.\ $2 ( 1 +
n_{\mathrm{RBF}} + n_{\varepsilon} ) = n$.

\begin{figure}[t!]
  \centering
  \includegraphics[scale=0.9]{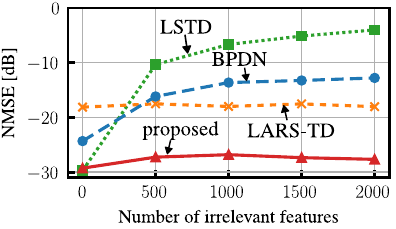}
  \caption{NMSE vs.\ number of irrelevant features for $1000$ samples.}
  \label{fig:demo_chainwalk_vallue_NMSE_noise}
  \vspace{-1ex}
\end{figure}

\begin{figure}[t!]
  \centering
  \includegraphics[scale=0.9]{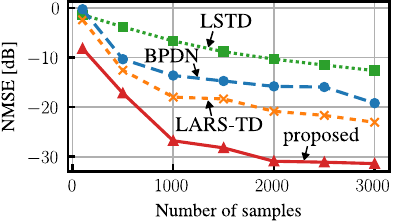}
  \caption{NMSE vs.\ number of samples for $1000$ irrelevant features.}
  \label{fig:demo_chainwalk_vallue_NMSE_samples}
\end{figure}

The proposed method is compared against LSTD~\cite{lagoudakis2003least}, LARS-TD~\cite{kolter2009regularization}, and
the state-of-the-art basis pursuit denoising (BPDN) approach~\cite{qin2014sparse}. The evaluation metric is the
normalized mean square error (NMSE) between the true value function and its estimate, defined as $\sum_{s=1}^{20}
[V^{*}(s) - \min_{a \in \mathfrak{A}} \hat{Q}(s, a)]^2 / \sum_{s=1}^{20} (V^{*}(s))^2$, where $\hat{Q}$ is an estimate of $Q^*$.  All
hyperparameters are carefully tuned for each method to achieve best performance. For LARS-TD, a small amount of
$\ell_2$-norm regularization is included when convergence cannot otherwise be guaranteed, as recommended
in~\cite{kolter2009regularization}.

Figures~\ref{fig:demo_chainwalk_vallue_NMSE_noise} and~\ref{fig:demo_chainwalk_vallue_NMSE_samples} depict NMSE as a
function of the number of irrelevant features and the number of samples, respectively. The reported results are averaged
over $30$ trials. While LARS-TD and BPDN yield more accurate value-function estimates than LSTD in the presence of many
irrelevant features, their performance remains limited due to the estimation bias induced by the $\ell_1$-norm
penalty. In contrast, the proposed method significantly outperforms all competitors by effectively mitigating this
bias. Furthermore, it maintains excellent accuracy even when the number of samples is small.

\section{Conclusions}

This paper introduced an effective batch algorithm for feature selection in RL, equipped with formal convergence
guarantees. A nonconvex PMC penalty was incorporated into the policy-evaluation objective to promote sparsity in the
feature-selection vector. The resulting optimization problem was reformulated as a nonmonotone inclusion, and the FRBS
method was applied with newly established convergence guarantees for a general class of nonmonotone inclusion problems
that subsumed the present setting. Numerical experiments showed that the proposed method substantially outperforms
state-of-the-art approaches in value-function estimation, achieving significantly reduced bias. Future research
directions include (i) extending the method to online learning and (ii) integrating it with the variational framework
of~\cite{akiyama2024nonparametric} for Q-function approximation in reproducing kernel Hilbert spaces (RKHS).

\clearpage
\bibliographystyle{IEEEtran}
\bibliography{draft_bib}

\end{document}